\documentclass[conference]{IEEEtran}

\usepackage{tabularx}
\usepackage{array}

\usepackage{amsmath,amsfonts,bm,bbm,enumitem}
\allowdisplaybreaks
\usepackage[mathscr]{euscript}
\usepackage{color}
\usepackage{amsthm}
\usepackage{graphicx}
\usepackage{url}
\usepackage{algorithm,algorithmic,multirow,hhline}
\usepackage{dblfloatfix}
\usepackage{amssymb}
\usepackage{mathalfa} 
\usepackage{caption}
\usepackage{comment}
\usepackage{subfigure}
\usepackage{pstricks}
\usepackage{subfloat}
\usepackage{setspace}
\usepackage{hyperref}
\usepackage{color}
\usepackage{booktabs,multirow}
\usepackage{enumitem}
\usepackage{multicol}
\usepackage{mathtools}

\usepackage{cite}

\makeatletter

\newcommand{\Rmnum}[1]{\expandafter\@slowromancap\romannumeral #1@}
\makeatother

\newtheorem{remark}{Remark}

\providecommand{\propositionname}{Proposition}

\usepackage[T1]{fontenc}
\usepackage{etoolbox}

\title{Energy-Efficient Model Compression and Splitting for Collaborative Inference 
Over Time-Varying Channels}

\author{\IEEEauthorblockN{Mounssif Krouka, Anis Elgabli, Chaouki Ben Issaid and Mehdi Bennis}
\IEEEauthorblockA{Centre for Wireless Communications (CWC), University of Oulu, 90014 Oulu, Finland,\\
Email: \{mounssif.krouka, anis.elgabli, chaouki.benissaid, mehdi.bennis\}@oulu.fi.}}

\begin{document}

\maketitle

\begin{abstract}
Today's intelligent applications can achieve high performance accuracy using machine learning (ML) techniques, such as deep neural networks (DNNs). Traditionally, in a remote DNN inference problem, an edge device transmits raw data to a remote node that performs the inference task. However, this may incur high transmission energy costs and puts data privacy at risk. In this paper, we propose a technique to reduce the total energy bill at the edge device by utilizing model compression and time-varying model split between the edge and remote nodes. The time-varying representation accounts for time-varying channels and can significantly reduce the total energy at the edge device while maintaining high accuracy (low loss). We implement our approach in an image classification task using the MNIST dataset, and the system environment is simulated as a trajectory navigation scenario to emulate different channel conditions. Numerical simulations show that our proposed solution results in minimal energy consumption and $CO_2$ emission compared to the considered baselines while exhibiting robust performance across different channel conditions and bandwidth regime choices.
\end{abstract}

\section{Introduction}

Recent advances in artificial intelligence (AI) have greatly led to developing new avenues for intelligent applications such as self-driving cars, social media, and proactive healthcare management \cite{AI, reconciling,abdel2020vehicular}. To achieve high accuracy, these applications require complex machine learning (ML) techniques such as deep neural networks (DNN) being the most widely used \cite{tariq2019speculative, Google}. 
\begin{figure}[t]
\centering
\includegraphics[width=9cm,height=10cm,keepaspectratio]{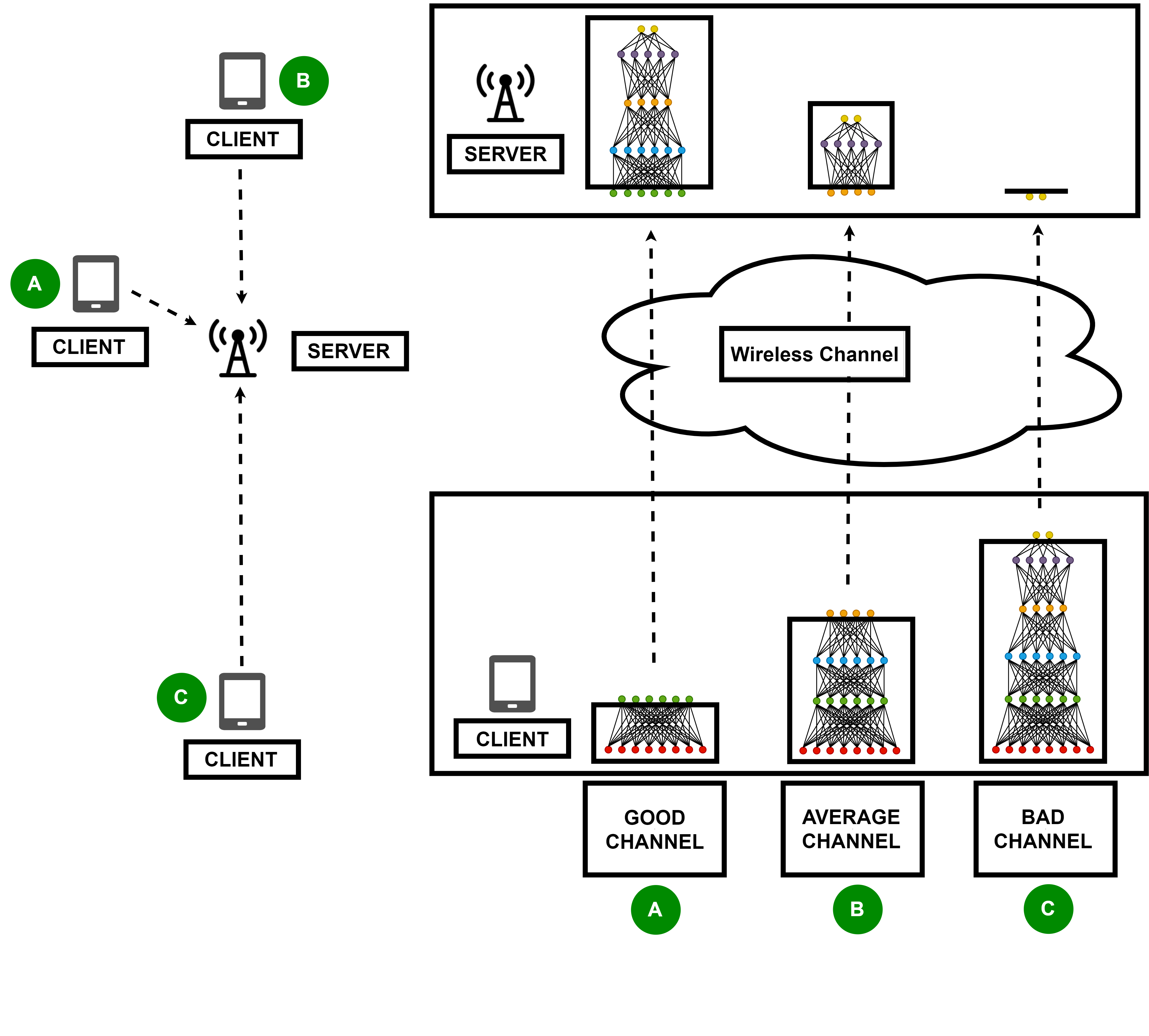}
  \caption{Remote inference system model where the edge device (client) sends a time-varying communication-efficient data representation to the remote node (server).}
\label{model}
\end{figure}

The traditional way to perform remote ML inference is to let the node exposed to the input data ({\it client}) transmit its data to a remote inferring node ({\it server}), usually through a wireless link where the intensive DNN computations are performed. However, the large size of the data to be shared as well as the privacy constraint limit the feasibility of this approach in practice \cite{privacy}.  Recently, embedded devices equipped with more enhanced computation capabilities allow for more effective data processing at the network edge \cite{liu2019performance, edge}. However, for large-size models and under edge devices' energy constraints, the inference task may not be locally performed. One way to minimize the energy bill of the client while ensuring high inference accuracy at the server is to split the model between both nodes. In this paper, we propose a novel approach that addresses the problem of minimizing the total energy (local computation and transmission energies) by: $(i)$ sparsifying the model while ensuring low loss (high accuracy) and $(ii)$ introducing a time-varying model splitting strategy between the client and the server. Formally, we consider the problem of minimizing the total energy under time-varying wireless channels while maintaining high accuracy (low loss).  

To tackle this problem, some recent works \cite{neuro, distributed, dynamic, autodidactic} propose task offloading to reduce the latency and throughput where the DNN is split into two parts, one at the client and one at the server. For example, in \cite{neuro},  model splitting is performed based on per-layer energy consumption and latency, and the work in \cite{distributed} suggests a technique to divide a DNN into multiple partitions based on a matching game theoretic approach. Authors in \cite{Jankowski_2020, shi2019improving} consider both model splitting and model \textit{compression} in order to improve the device computation efficiency. \textit{Compression} is performed by removing some weights/biases. In \cite{Jankowski_2020}, authors propose a joint feature compression and transmission scheme for efficient inference tasks, for different network splitting points. The work in \cite{shi2019improving} puts forward a 2-step pruning framework for DNN partitioning between mobile devices and edge servers.
Nevertheless, none of aforementioned works combine model \textit{compression} with the dynamic effect of the instantaneous time-varying channel gain on the DNN splitting strategy.

In this paper, we analyse different aspects of energy and attempt to minimize the total energy at the client-side by performing model \textit{compression} in order to reduce the model size and time-varying model splitting between the client and the server to account for the trade-off between local computation and transmission energies over time-varying wireless channels. Note that the output of each layer in the DNN is considered as one representation for the input data; hence, as shown in Fig. \ref{model}, at any given time, we decide which representation to send such that the total computation energy to produce that representation plus the total communication energy of this specific representation is minimized for the given channel state. Our simulation results show that our proposed approach significantly reduces the total energy bill of the client as well as its corresponding $CO_2$ emission, and achieves better performance compared to several baselines under different channel gains and bandwidth regimes.

The rest of the paper is organized as follows. In section \ref{sys}, we describe the system model and the problem formulation. In section \ref{proposed}, we describe our proposed solution. Before concluding, we present the simulations results and discussion in section \ref{results}. 
\section{System model and problem formulation} \label{sys}
In this work, we study a system that comprises a client transmitting its data representation to a remote server for a machine learning (ML) inference task. The client's goal is to operationalize a trained ML model and send a communication-efficient representation to the server, as depicted in Fig. \ref{model}. Our objective is to minimize the client's energy bill over a time-varying wireless channel while maintaining high inference accuracy. In this work, we consider the computation energy $(E_{c})$, memory access energy $(E_{m})$, and transmission energy $(E_{tr})$. To formulate the problem, we first need to quantify the different energy aspects.
\subsection{Computation Energy}
The computation energy $(E_{c})$ is the required energy for performing arithmetic operations such as addition and multiplication. We denote by $e_M$ and $e_A$ the energy for a single multiplication and addition operations, respectively. In Table \ref{Table:COMP_ENERG}, we report the energy values of these operations based on the benchmark \cite{Memory_access}. As a consequence, the computation energy can be computed as follows
\begin{align}\label{eq:E_c}
E_{c} = e_M \times M + e_A \times A,
\end{align}
where $A$ and $M$ are the required number of addition and multiplication operations, respectively.
\begin{table}[t]
\vskip 0.15in
\begin{center}
\begin{small}
\begin{sc}
\begin{tabular}{lcc}
\toprule
Arithmetic operation  & ADD & MUL \\
\midrule
8-bit Integer   &     0.03 pJ&  0.2 pJ\\
16-bit Floating point &   0.4 pJ  &    1.1 pJ \\
32-bit Integer  &    0.1 pJ &    3.1 pJ\\
32-bit Floating point  &    0.9 pJ &    3.7 pJ \\
\bottomrule
\end{tabular}
\caption{Approximate energy costs for different arithmetic operations in 45nm 0.9V.} 
\label{Table:COMP_ENERG} 
\end{sc}
\end{small}
\end{center}
\vskip -0.1in
\end{table}
\begin{table}[t]
\vskip 0.15in
\begin{center}
\begin{small}
\begin{sc}
\begin{tabular}{lc}
\toprule
Memory size & 64-bit Memory access \\
\midrule
8KB                             & 10 pJ \\
32KB                            & 20 pJ  \\
1 MB                            & 100 pJ \\
DRAM                            & 1.3-2.6 nJ  \\
\bottomrule
\end{tabular}
\caption{Memory access energy expenditure (consumption) in 45nm 0.9V.}
\label{Table:MEMORY_ENERG}  
\end{sc}
\end{small}
\end{center}
\vskip -0.1in
\end{table}
\subsection{Memory Access Energy}
The memory access energy $(E_{m})$ is the energy utilized to fetch data from memory (cache or RAM) and move the necessary data to the arithmetic/logic unit. From Table \ref{Table:MEMORY_ENERG}, we notice that the energy cost of the DRAM access is much higher than any computation operation. This is due to the required static power to keep the I/O active \cite{Memory_access}. Therefore, the memory access energy is given as follows 
\begin{align}\label{eq:E_m}
E_{m} = \Gamma \times e_m,
\end{align} 
where $e_m$ is the energy for fetching/decoding one element  from memory, and $\Gamma$ is the needed number of elements (weights and biases in the case of DNN).
\begin{figure*}[t]
\centering
\includegraphics[scale=0.4]{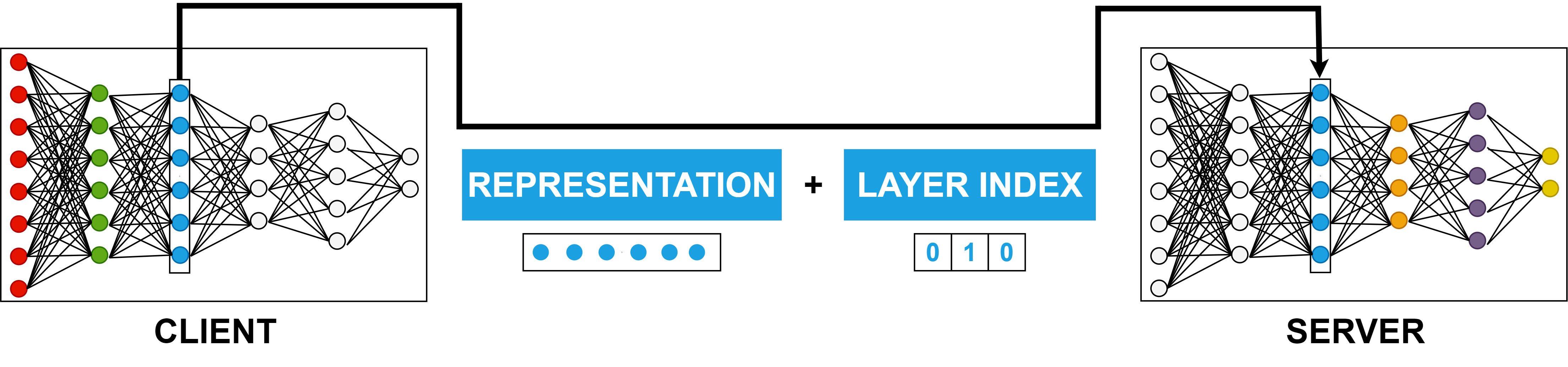}
  \caption{Illustration of the proposed approach.}
\label{index}
\end{figure*}
\subsection{Transmission Energy}
The transmission energy $(E_{tr})$ is the energy used for transmitting the data representation to the server over the wireless channel. This energy depends on both the size of the transmitted representation as well as the channel condition. For the transmission energy, we consider that the output of each neuron in the cut layer (representation) is transmitted using 32 bits (full-precision communication). We let $W$ and $h(t)$ be the bandwidth and the flat fading channel gain at time instant $t$, respectively. The channel model is generated according to a Rayleigh fading with zero mean and unit variance, i.e. $h \sim \mathcal{C N}(0,\,1)$. Thus, according to Shannon's formula, the asymptotic transmission rate at time $t$ is given by
\begin{align}\label{eq:rate}
R(t) = W \log_2\left(1 + \frac{P(t) |h(t)|^2}{N_0W}\right),
\end{align} 
where $N_0$ and $P(t)$ are the power spectral density and the transmission power of the client, respectively.

The time $\tau$ required to transmit $d$ elements should satisfy the following inequality 
\begin{align}\label{eq:ineq}
\int_{t=0}^{\tau} R(t)dt \geq 32d,
\end{align}
where $\tau$ and $d$ are the uploading time and the number of transmitted elements, respectively. Hence, the transmission energy at the client side can be written as
\begin{align}
E_{tr} = \tau \int_{t=0}^{\tau} P(t) dt.
\end{align} 
\begin{remark}
Throughout this paper, we assume that we have a perfect estimation of $h(t)$ for the period $t \in [0, \tau)$.
\end{remark}
As a consequence, the total energy budget of the client per inference can be expressed as follows
\begin{align}\label{eq:E}
\nonumber E &= E_{c} + E_{m} + E_{tr}\\
 &= E_p + E_{tr},
\end{align}
where the processing energy $E_{p}$ is the sum of the computation and memory access energies. In the remainder of this paper, we add the subscript $i$ to the energy definition to account for the fact that the energy depends on the channel at the $i^{th}$ inference time.
\subsection{Problem Formulation}
We aim to minimize the client's total energy bill while maintaining high accuracy. Therefore, we can formulate the following problem 
\begin{align} \label{eq:1}
   \underset{\boldsymbol{\theta}_{c,i}, \boldsymbol{\theta}_{s,i}}{\min} \,\,\,\,\,& \lambda \cdot loss(\boldsymbol{\theta}_{c,i}, \boldsymbol{\theta}_{s,i}) +  (1 - \lambda) \cdot  \lim\limits_{\mathrm{I} \to \infty}\frac{1}{\mathrm{I}} \sum_{i=1}^{\mathrm{I}} E_i(\boldsymbol{\theta}_{c,i}), 
\end{align}
where $loss$ is the training loss function, $\boldsymbol{\theta}_{c,i}$ and $\boldsymbol{\theta}_{s,i}$ are the client and server models at the $i^{th}$ inference time, respectively, $\mathrm{I}$ is the total number of inferences, and $\lambda \in (0, 1)$ is a parameter that controls the trade-off between achieving low loss (high accuracy) and low energy consumption at the client's side.

Since the $loss$ function does not depend on the choice of the cut layer (transmitted representation), problem \eqref{eq:1} can be re-written as
\begin{align} \label{eq:001}
   \underset{\boldsymbol{\theta}_{c,i}, \boldsymbol{\theta}_{s,i}}{\min} \,\,\,\,\,& \lambda \cdot loss(\boldsymbol{\theta}) + (1 - \lambda) \cdot  \lim\limits_{\mathrm{I} \to \infty}\frac{1}{\mathrm{I}} \sum_{i=1}^{\mathrm{I}} E_i(\boldsymbol{\theta}_{c,i}), 
\end{align}
where $\boldsymbol{\theta} = [\boldsymbol{\theta}_{c,i}, \boldsymbol{\theta}_{s,i}]$ is the whole model, which is constant across different transmissions.

Problem \eqref{eq:001} is hard to solve due to the lack of channel prediction for the whole period and non-convexity of the feasible set. In fact, the cut layer index belongs to a discrete set, which makes problem \eqref{eq:001} a combinatorial one. Next, we propose an approximate and a low-complexity approach to solve problem \eqref{eq:001}.
\begin{figure*}[t]
\includegraphics[width=\textwidth]{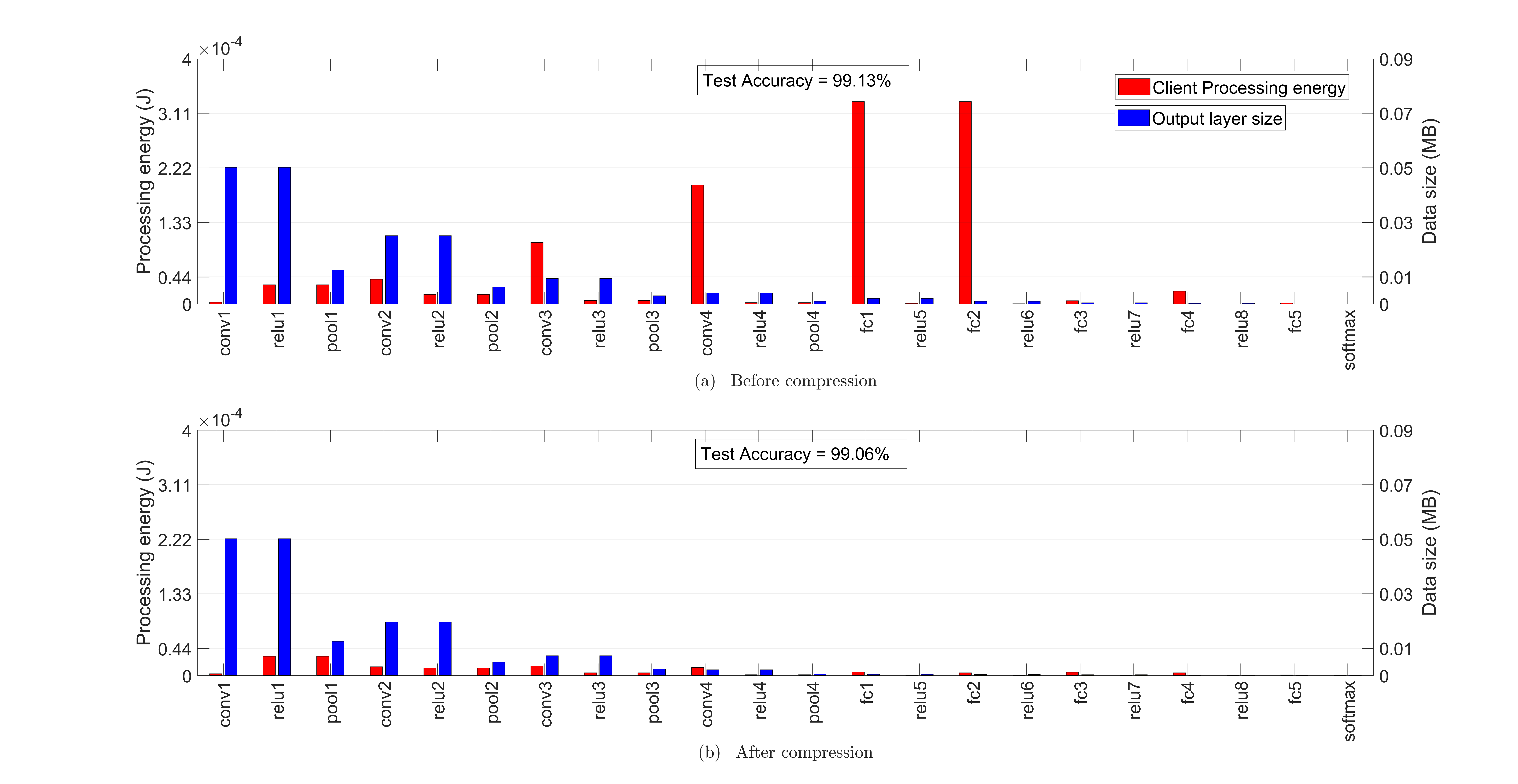}
  \caption{Per layer processing energy and output size before and after compression.}
\label{per_layer}
\end{figure*}
\section{Proposed algorithm} \label{proposed}
The energy minimization problem formulated in \eqref{eq:001} is controlled by two system aspects, namely the neural network (NN) complexity and the channel observation. We note that $E_{p}$ depends solely on the number of weights and biases. This implies that lowering the model complexity reduces the processing energy. Accordingly, one solution to reduce the NN size is to perform model \textit{compression} in the training phase.

On the other hand, the transmission energy depends on the size of the transmitted representation (number of neurons) as well as on the channel observation. Consequently, in addition to model complexity reduction, the proper selection of the layer over which the representation is transmitted contributes to minimizing the total energy. 
Therefore, our proposed solution mandates decomposing the original problem into two sub-problems: $(i)$ model compression and $(ii)$ representation selection.
\subsection{Model Compression}\label{Model_compression}
Model \textit{compression} leads to minimizing both the communication and computation energies since it sparsifies the model and also reduces the transmitted representation size. However, to sparsify the model while maintaining low loss/high accuracy, one can formulate the following sub-problem 
\begin{align}\label{eq:comp1}
 \underset{\boldsymbol{\theta}} {\min} ~~ & \mu \cdot loss(\boldsymbol{\theta}) + (1-\mu) \cdot \sum_{l=1}^{L}{\lVert \theta_l \rVert}_0,
\end{align}
where $L$ is the number of layers in the NN, $\theta_l$ is the model of the $l^{th}$ layer, and $\mu$ is a parameter that controls the trade-off between model sparsification and the \textit{loss} minimization.
Note that problem \eqref{eq:comp1} is hard to solve since the $\ell_0$ norm minimization is a well-known NP-hard problem. Thus, instead of solving problem \eqref{eq:comp1}, we propose the following two-steps procedure
\begin{enumerate}
\item Solve the convex relaxation of the $\ell_0$ norm minimization as follows
\begin{align}\label{eq:comp2}
 \underset{\boldsymbol{\theta}} {\min} ~~ & \mu \cdot loss(\boldsymbol{\theta}) + (1-\mu) \cdot \sum_{l=1}^{L}{\lVert \theta_l \rVert}_1.
\end{align}
\item Model \textit{pruning} by eliminating all the parameters with values below a predefined pruning threshold.  
\end{enumerate}
\begin{remark}
Note that problem \eqref{eq:comp2} is solved once since its solution does not depend on the time-varying channel.
\end{remark}
\begin{remark}
Practically, $\mu$ is chosen such that $(1-\mu) << 1$. This choice of $\mu$ can lead to model sparsification at a low cost in terms of accuracy drop.
\end{remark}
\subsection{Representation Selection}
Solving the sub-problem introduced in Sec. \ref{Model_compression} results in a sparse version of the original model. Nevertheless, it does not provide information on which representation to send such that the communication energy is minimized. Selecting the appropriate representation, i.e., the output of which layer, to transmit is crucial in reducing the transmission energy for the given channel observation. The selection of the representation to be transmitted has a direct effect on the second term in problem \eqref{eq:001}, i.e, the total energy.

As depicted in Fig. \ref{index}, we consider that the remote server holds a copy of the client's model. When the client transmits a particular layer's output (representation), this layer's index is also shared, so the server can identify which layer to start from and carry on the inference task. Our proposed solution suggests that we can maintain a trade-off between the processing energy and the transmission energy by selecting the optimal layer $l \in \{1, \dots, L\}$ for a given channel observation. Formally, we choose the output of layer $l$ at the $i^{th}$ transmission, where $l$ is the layer's index that minimizes $E_i(\bm{\theta}_{c,i})$. 
\section{Simulation Results} \label{results}
\begin{figure*}[t]
\hspace*{0.4cm}
\includegraphics[width=\textwidth-1cm,keepaspectratio]{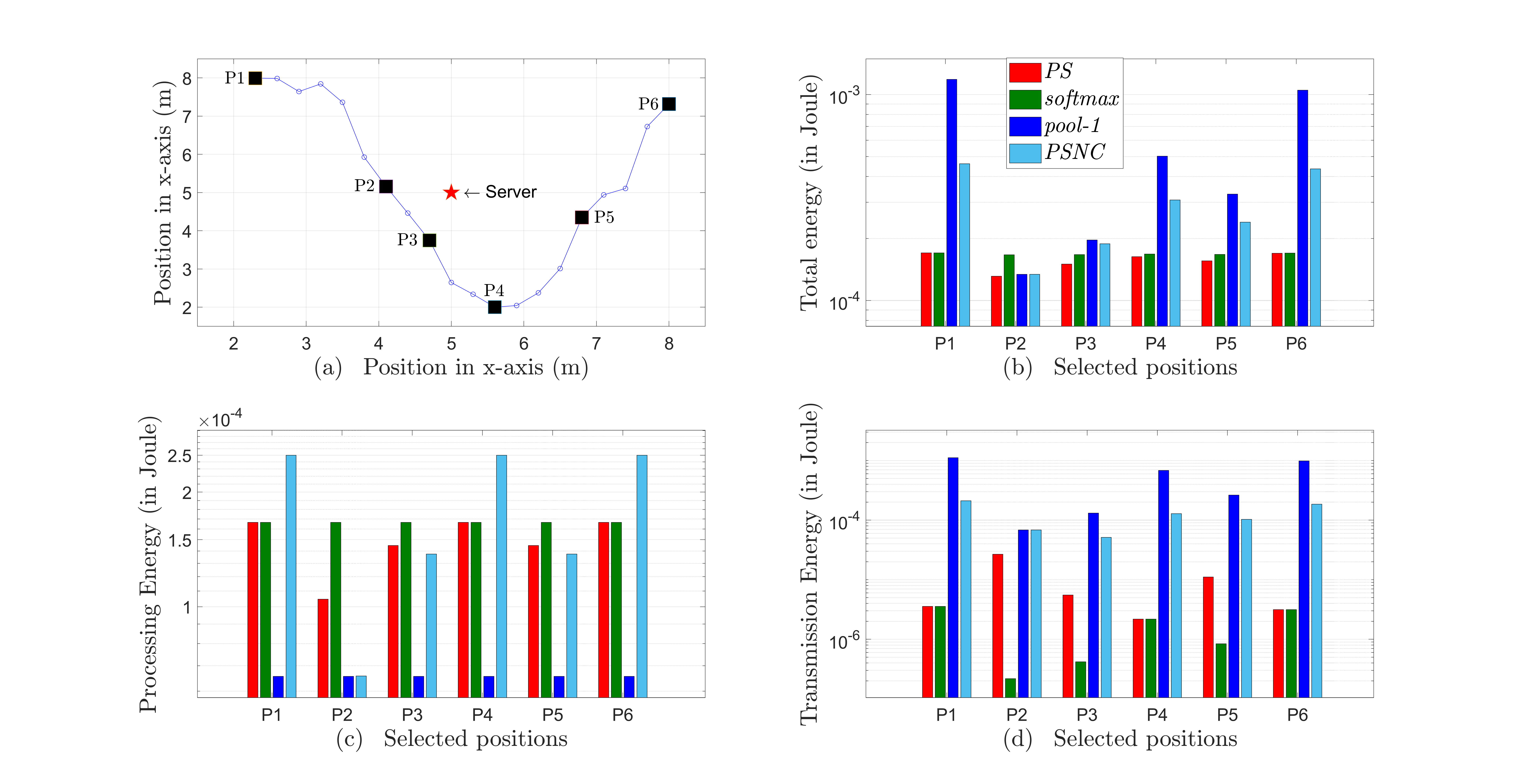}
  \caption{Cut layer selection for different channel observations and its corresponding energy consumptions.}          
\label{channel_conditions}
\end{figure*}
For our simulations, we consider the image classification task where the MNIST dataset is used \cite{MNIST}. The dataset consists of a set of $28 \times 28$ gray-scale images with $60K$ for training and $10K$ for testing. The images represent handwritten digits ranging from $0$ to $9$. For our NN model, we use the convolutional NN (CNN) architecture. Our original model consists of $4$ convolutional layers (\textit{conv}) with a $5 \times 5$ filter and $16$, $32$, $48$, and $64$ channels, respectively. Each \textit{conv} layer is followed by a \emph{Relu} activation function and a pooling layer (\textit{pool}) with \textit{stride} $2$ for size reduction. After that, we have $5$ fully connected layers (\textit{fc}) with $512$, $256$, $128$, $64$, and $10$ neurons, respectively. The \emph{Relu} activation function is applied to all \textit{fc} layers expect for the last layer where \textit{softmax} function is applied.  We use $P = 1 mW$, $N_0 = 10^{-9}$, batch size = $64$, learning rate $\alpha = 10^{-3}$, $\mu = 0.9999$, and $W = 5 MHz$.

\begin{figure}[t]
\centering
\includegraphics[width=8cm,height=7cm,keepaspectratio]{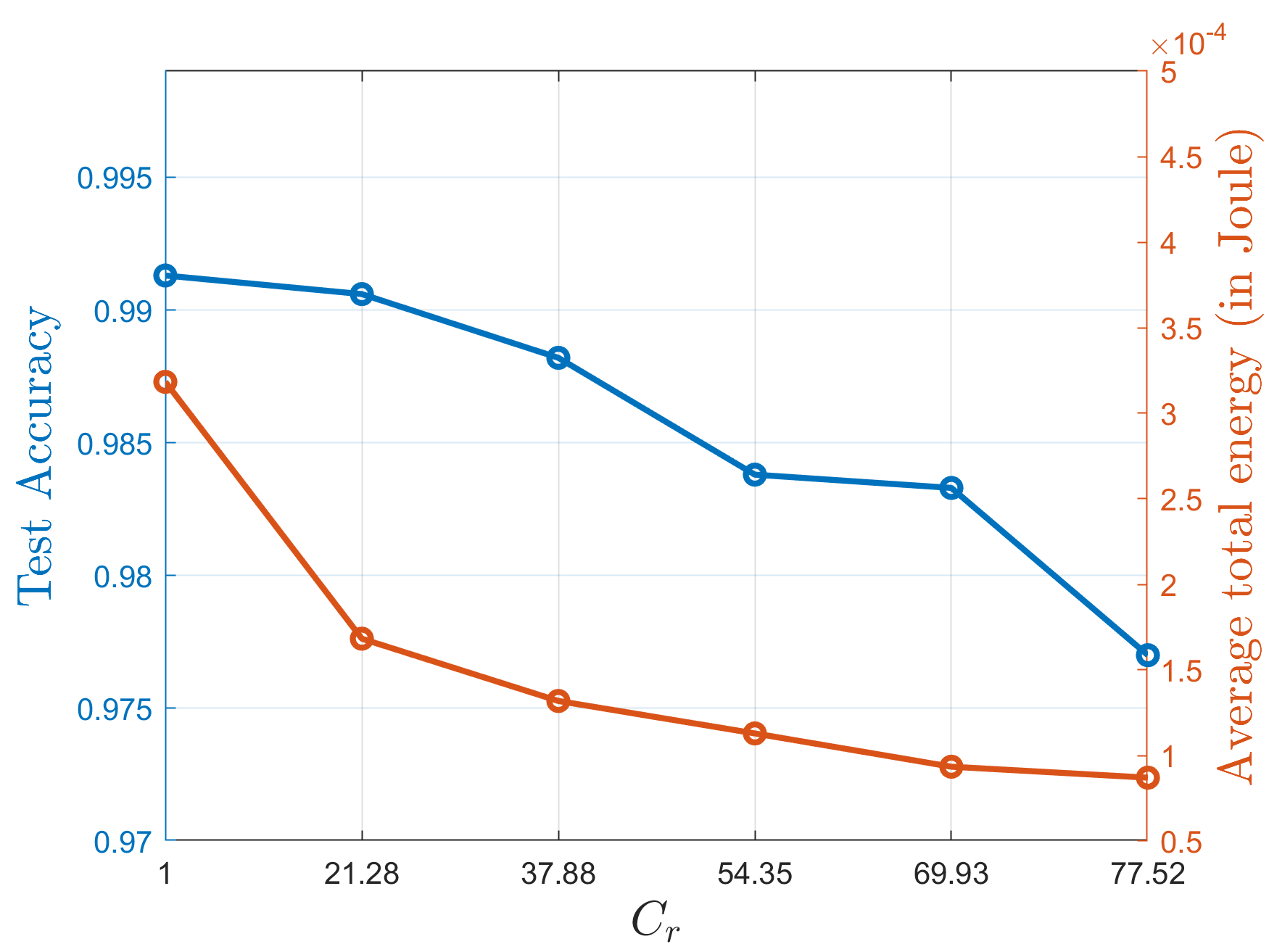}
  \caption{Average total energy and test accuracy values for different compression ratios.}
\label{acc_en}
\end{figure} 

In order to estimate the equivalent $CO_2$ emissions of the client, we use the data reported in \cite{CO2}, for different European Union (EU) countries in terms of carbon index (CI), where CI represents the greenhouse gas emission intensity calculated as kg of CO2 emissions per kWh $(kg \, CO_{2}e/kWh)$ from public electricity production. In particular, we consider the average EU case with $CI = 0.275$.

Fig. \ref{per_layer} shows the processing energy and the output size for every layer in the NN before and after compression. We notice that the compression step significantly reduces the processing energy for more than $98\%$ for both \textit{fc1} and \textit{fc2} layers which comprise around $60\%$ of the NN parameters. Moreover, the decrease in the output size of the layers is reflected by eliminating entire neurons after pruning.   

In Fig. \ref{acc_en}, we plot the average total energy of the system and the testing accuracy for different values of the compression ratio $C_r$. The compression ratio is a measure of the relative reduction in the number of model parameters defined as the ratio of the total number of non-zero parameters of the original model $\bar{\theta^o}$ with respect to the number of parameters of the compressed model $\bar{\theta^c}$. As the compression ratio increases, more energy expenditure can be saved, but at the cost of decreasing test accuracy. This is due to the fact that aggressive sparsification leads to eliminating more parameters which may decrease the accuracy.

Fig. \ref{channel_conditions} shows the performance of the proposed layer selection (\textit{PS}) compared to the following baselines
\begin{itemize}
\item \textit{Pool-1}: The client processes the data up to the first pooling layer. The pooling operation is used for dimensionality reduction of the feature maps.
\item \textit{softmax}: The client runs over all the model layers and transmits the labels.
\item  \textit{PSNC}: The client selects the optimal layer for a given channel observation employing the original model (without compression).
\end{itemize}

\begin{table*}[t]
\vskip 0.15in
\begin{center}
\begin{small}
\begin{sc}
\begin{tabular}{lcccc}
\toprule
& pool-1 & softmax & PS & PSNC\\
\midrule
&  &  $W = 15 KHz$ &  &  \\
\midrule
$E~~(J)$  & $8.66 \times 10^{-4}$  & $1.69 \times 10^{-4}$  & $1.69 \times 10^{-4}$  & $4.36 \times 10^{-4}$\\
$E_p~~(J)$ & $0.66 \times 10^{-4}$  & $1.67 \times 10^{-4}$  & $1.64 \times 10^{-4}$  & $2.60 \times 10^{-4}$\\
$E_{tr}~~(J)$ & $7.98 \times 10^{-4}$  & $0.03 \times 10^{-4}$  & $0.05 \times 10^{-4}$  & $1.76 \times 10^{-4}$  \\ 
$EU~~CO_{2}e~~(kg)$ & $0.66 \times 10^{-10}$ & $0.13 \times 10^{-10}$ & $0.13 \times 10^{-10}$ & $0.33 \times 10^{-10}$ \\
\midrule
&  &  $W = 15 MHz$ &  &  \\
\midrule
$E~~(J)$  & $1.24 \times 10^{-4}$  & $1.67 \times 10^{-4}$  & $0.53 \times 10^{-4}$  & $0.61 \times 10^{-4}$  \\
$E_p~~(J)$  & $0.66 \times 10^{-4}$  & $1.67 \times 10^{-4}$  & $0.23 \times 10^{-4}$  & $0.23 \times 10^{-4}$  \\
$E_{tr}~~(J)$ & $0.58 \times 10^{-4}$  & $0.002 \times 10^{-4}$ & $0.31 \times 10^{-4}$  & $0.38 \times 10^{-4}$ \\
$EU~~CO_{2}e~~(kg)$ & $0.1 \times 10^{-10}$  & $0.12 \times 10^{-10}$ & $0.04 \times 10^{-10}$ & $0.05 \times 10^{-10}$ \\
\bottomrule
\end{tabular}
\caption {Simulation results for different bandwidth values and baselines.} 
\label{Table:BW} 
\end{sc}
\end{small}
\end{center}
\vskip -0.1in
\end{table*}

Fig. \ref{channel_conditions}(a) depicts a trajectory of the client in a 2-dimensional grid to simulate different channel conditions. In Fig. \ref{channel_conditions}(b), we plot the total energy for the different selected locations. We notice that \textit{PS} outperforms all the baselines. \textit{pool-1} has the worst performance since the client is far from the server, and the transmit energy is the dominant part. On the other hand, when the channel is good (at P2), \textit{softmax} performance is bad when processing energy is dominant. In Fig. \ref{channel_conditions}(c), processing energy is plotted. We notice that \textit{pool-1} performs better than all the other cases at the cost of consuming the highest transmission energy because of the large output size for \textit{conv} layers, as seen in Fig. \ref{channel_conditions}(d). The baseline \textit{PSNC} uses up more processing energy seeing the high complexity of the original model. Fig. \ref{channel_conditions}(d) illustrates the transmission energy results. We observe that \textit{softmax} is always better than the other cases due to the small size of the labels, at the price of running through all the layers of the NN.

Table \ref{Table:BW} contains different simulation results for different values of the bandwidth. We observe that \textit{PS} outperforms all the other baselines in terms of total energy consumption and consequently $CO_2$ emission. For $W = 15KHz$, \textit{softmax} performs slightly similar to \textit{SP}. This is consequent to the high transmission energy at the low bandwidth regime. For $W = 15MHz$, the abundance of bandwidth enables the client to process few layers and transmit larger output sizes in order to save processing energy. This is observed by the remarkable performance of \textit{PSNC} since most NN parameters are positioned in the \textit{fc} layers.

\section{Conclusion} \label{conclusion}
In this paper, we tackle the problem of collaborative inference where the goal is to reduce the total energy bill at the edge device by utilizing model compression and time-varying model split between the edge and remote nodes. Numerical results show that our proposed scheme outperforms the considered baselines in terms of the total energy consumption and $CO_2$ emission. In particular, our scheme maintains a robust performance for different channel conditions and  bandwidth regime choices. For future work, we will 
extend our current system model to the multi-user case and investigate system latency and reliability under finite block-length regime.
\bibliographystyle{IEEEtran}
\bibliography{references}
\end{document}